\documentclass[runningheads]{llncs}

\usepackage{subcaption}
\usepackage{graphicx}
\usepackage{caption}
\usepackage{amsmath,amsfonts,amssymb,bbm}

\usepackage{hyperref}

\usepackage{tabularx}
\usepackage{multirow}

\usepackage{graphicx}
\begin{document}
\title{Regularization, Semi-supervision, and Supervision for a Plausible Attention-Based Explanation \thanks{Work partially funded by grant ANR-19-CE38-0011-03 from the French national research agency (ANR).}}
\titlerunning{Towards a Plausible Attention Explanation}
%
\author{Duc Hau Nguyen\inst{1}\orcidID{0000-0002-4061-3114} \and
Cyrielle Mallart\inst{3}\orcidID{0009-0001-1420-9548} \and
Guillaume Gravier\inst{2}\orcidID{} \and
Pascale Sébillot\inst{1}\orcidID{0000-0002-5429-4302}
}
\authorrunning{D.H. Nguyen et al.}
%
\institute{Univ Rennes, CNRS, Inria, INSA Rennes - IRISA, Rennes, France, \email{\{duc-hau.nguyen,pascale.sebillot\}@irisa.fr}
\and
Univ Rennes, CNRS, Inria - IRISA, Rennes, France, \email{guillaume.gravier@irisa.fr}\\
 \and
 University of Rennes 2, Rennes, France, \email{cyrielle.mallart@univ-rennes2.fr}
}
\maketitle              
\begin{abstract} Attention mechanism is contributing to the majority of recent advances in machine learning for natural language processing. Additionally, it results in an attention map that shows the proportional influence of each input in its decision. Empirical studies postulate that attention maps can be provided as an explanation for model output. However, it is still questionable to ask whether this explanation helps regular people to understand and accept the model output (the plausibility of the explanation). Recent studies show that attention weights in the RNN encoders are hardly plausible because they spread on input tokens. We thus propose 3 additional constraints to the learning objective function to improve the plausibility of the attention map: regularization to increase the attention weight sparsity, semi-supervision to supervise the map by a heuristic and supervision by human annotation. Results show that all techniques can improve the attention map plausibility at some level. We also observe that specific instructions for human annotation might have a negative effect on classification performance. Beyond the attention map, the result of experiments on text classification tasks also shows that no matter how the constraint brings the gain, the contextualization layer plays a crucial role in finding the right space for finding plausible tokens.

\end{abstract}
\section{Introduction}

Attention mechanisms \cite{luong_EffectiveApproachesAttentionbased_2015} play a crucial role in recent success across many natural language processing (NLP) tasks and are present in most recent neural models. As a layer in a complex neural network, the mechanism attributes weight to each input token and encodes by weighted summing the input vectors into a context vector. When this context vector is used for prediction, the weight vector, also called \textit{attention map}~\cite{jacovi_FaithfullyInterpretableNLP_2020}, can be considered as an explanation by showing the degree of influence of each input token in the prediction.

Despite the potential of attention maps as a form of explanation, there are concerns~\cite{jacovi_FaithfullyInterpretableNLP_2020} about their validity on two properties that are not guaranteed: faithfulness and plausibility. Faithfulness, a widely discussed problem \cite{mohankumar_TransparentExplainableAttention_2020,bibal_AttentionExplanationIntroduction_2022}, focuses on whether the weight associated with a token reflects its influence on the prediction. Plausibility refers to the extent to which the attention map can resemble human reasoning \cite{mohankumar_TransparentExplainableAttention_2020,wiegreffe_AttentionNotNot_2019}. 

While plausibility is an interesting feature that allows to present an easily comprehensible way to individuals with limited knowledge of neural models without additional computational costs, the contributions in this direction remain limited and rare. Given that multiple studies have suggested that raw attention weights lack plausibility (see, e.g., \cite{nguyen_StudyPlausibilityAttention_2021}), the issue of forcing their plausibility is an obvious one that calls for further exploration. As it is proven possible to incorporate constraints on attention while maintaining satisfactory performance \cite{pruthi_LearningDeceiveAttentionBased_2020,jain_AttentionNotExplanation_2019,wiegreffe_AttentionNotNot_2019}, we propose three approaches for enforcing plausibility constraints on attention maps, namely, sparsity regularization, semi-supervised learning, and supervised learning.

The main contributions in this paper are : (1) we can to some extent force the model plausibility (as demonstrated by supervision) at no accuracy cost, (2) both regularization and semi-supervision can optimize the plausibility but the latter offers a solution without compromising performance and (3) the deep contextualization is harmful to attention plausibility. The last result provides insight into why this hardly transfers to transformers.

\section{Related Works}

The attention mechanism is widely used as a possible feature to explain the model decision~\cite{ousidhoum_ProbingToxicContent_2021,sun_UnderstandingAttentionText_2020}. However, the local explanation is facing an issue that the attention map can be manipulated while keeping the same prediction \cite{jain_AttentionNotExplanation_2019,wiegreffe_AttentionNotNot_2019,vashishth_AttentionInterpretabilityNLP_2019}. While this feature is considered as a weak faithful explanation \cite{serrano_AttentionInterpretable_2019,bai_WhyAttentionsMay_2021}, this enables the selection of a plausible map.

Among the few studies on attention supervision, \cite{mcguire_SentimentAnalysisCognitive_2021} showed that supervision can harm classification performance in sentiment classification tasks. Regularization was considered to circumvent the issue of a rather flat distribution of attention weights as reported by~\cite{jia_ARNORAttentionRegularization_2019}. \cite{mohankumar_TransparentExplainableAttention_2020} suggested an additional constraint in the learning objective to force this representation to be sparse.

To overcome the lack of human references in many datasets, many contributions offer task-specific solutions, such as~\cite{nguyen_WhoKilledPolice_2018} guiding attention based on topic-related vocabulary and~\cite{lehman_InferringWhichMedical_2019} using a WordNet-based heuristic for evidence inference, while~\cite{nguyen_StudyPlausibilityAttention_2021} provides an effective heuristic map that is closer to human annotation but only for NLI. While the authors of existing techniques have not fully explored their effects and limitations in different tasks, this study aims to provide a comprehensive view of how different techniques improve attention plausibility. 

Hard attention, also referred to as rationalized learning in the literature \cite{chen_CanRationalizationImprove_2022}, is an alternative form of the attention mechanism that comprises two components: a generator function that masks irrelevant input tokens, and a predictor that is trained to make predictions on the remaining inputs. While hard-attention is advantageous with respect to soft-attention in robustness and faithfulness aspects, it introduced a trade-off between sparsity and accuracy because the full context is inaccessible \cite{paranjape_InformationBottleneckApproach_2020}. 

Other post-hoc explanation techniques can provide faithful explanations (such as gradient-based methods or feature suppression), they present two main drawbacks: (i) incurring additional computational costs during each inference and (ii) offering benefits only to the model developer, without the flexibility to impose constraints for plausible explanations \cite{bastings_ElephantInterpretabilityRoom_2020} while its explanation cannot be guaranteed to be plausible for end-users\cite{bibal_AttentionExplanationIntroduction_2022,nguyen_FiltrageRegularisationPour_2022}.

To the best of our knowledge, no existing study has brought a broad and comprehensive overview of how different techniques improve attention plausibility. Although regularization techniques are independent of human annotation and heuristics can overcome the lack of human annotation, it is still unclear how they improve plausibility compared to supervision. Furthermore, the authors of the existing techniques suggest improvement without questioning their implications and limitations in different tasks, especially in soft-attention models. This study focuses on addressing these fundamental issues and does not include a comparative analysis of hard-attention techniques and post-hoc explanation methods but they are promising for future works.

\section{Tasks and Datasets}
\label{sec:data}

To ensure the generalization of our findings across different tasks, we investigate three different datasets from the ERASER benchmark~\cite{deyoung_ERASERBenchmarkEvaluate_2020} and~\cite{wiegreffe_TeachMeExplain_2021} designed for plausibility studies.

The e-SNLI corpus \cite{camburu_ESNLINaturalLanguage_2018}, a reference dataset in natural language inference (NLI), consists of pairs of sentences, a premise and a hypothesis with a label stating whether the hypothesis entails, contradicts, or is unrelated to the premise. The annotators also answered the question \textit{Why is a pair of sentences in a relation of entailment, neutrality, or contradiction?} by highlighting the relevant words in both the premise and hypothesis and providing a short explanatory text. The corpus consists of 549,367 sentence pairs for training and 9,842 pairs for the validation and test sets respectively. Note that the SNLI corpus is known to have artifacts \cite{gururangan_AnnotationArtifactsNatural_2018}, where some lexical fields appear mostly in one class. Also, the annotation instruction in e-SNLI leads to some particularities, such as not highlighting the common words between premise and hypothesis, thus making the annotation not convincing in some cases. 

The HateXPlain dataset \cite{mathew_HateXplainBenchmarkDataset_2021} is conceived by gathering posts from social networks that were labeled for the detection of hate speech. Each post belongs to either one of three labels:  offensive, hateful, and normal speech. Annotators were also instructed to highlight the relevant part of the post to justify their choice of a label. Overall, the corpus consists of 15,383 posts for training, 1,922 for validation, and 1,924 for testing.

Yelp-Hat dataset \cite{sen_HumanAttentionMaps_2020a} is obtained by gathering reviews on restaurants from a website and by asking reviewers to highlight parts of the text to justify their choice. The corpus consists of 3,482 reviews for training and validation. \footnote{15 incoherent samples are excluded, such as incompatible annotation maps and number of tokens in reviews. Since we don't have a validation split, before experimentation Yelp-Hat is split randomly into 2,436 sentences for training and 1,046 for validation.}.

\section{Attention Mechanism on RNN Encoders}
\label{sec:model}

Being one of the most studied in NLP yet the most controversial in the explainability debate \cite{bibal_AttentionExplanationIntroduction_2022}, we employ thus the attention model in RNN encoders. Preliminary experiments on BERT-like self-attention models have shown little hope in finding a single layer or head to provide plausible explanations. 

\begin{figure}
     \vspace{-1em}
     \centering
     \includegraphics[width=\textwidth]{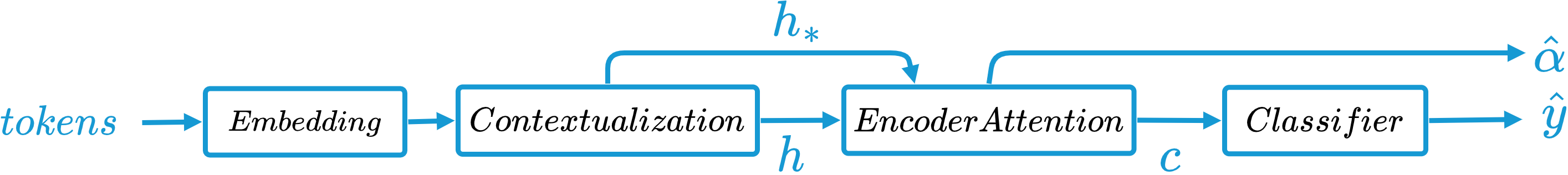}
     \caption{Generic architecture of a RNN-based attention model for classification.}
     \label{fig:architecture}
     \vspace{-2em}
\end{figure}

The model, illustrated in Figure~\ref{fig:architecture}, consists of an embedding layer and a bi-LSTM layer, which produce contextualized token representations $h_i$ for each token at position $i$ in a sentence of length $L$, as well as a sentence embedding $h_\ast$, which is the concatenation of the forward and backward last state. The attention encoder assigns weights $\hat{\alpha}_i$ to each $h_i$ and computes a context vector $c$ through a weighted sum. Finally, a multilayer perceptron classifier is applied to $c$ for prediction. We also consider attention weights on each input token in the loss function. To simplify notation, we use the notation $h = [h_1, ..., h_L]^\intercal = [h_i]_{i=1}^L $ to denote the sequence of bi-LSTM outputs and $\hat{\alpha} = [\hat{\alpha}_i]_{i=1}^L$ to denote the attention weights. In this paper, we distinguish the model attention map $\hat{\alpha}$ from $\alpha$ which refers to the human annotation binary map.

The attention layer is adapted differently for various tasks. We begin by describing the attention layer formally as a function that takes query $q$, key $k$, and value $v$ in input and produce $c$ and $\hat{\alpha}$ as output \cite{luong_EffectiveApproachesAttentionbased_2015}:

\begin{equation} 
    \label{eq:attention}
    c, \hat{\alpha} = Attention(q, k, v)
\end{equation}

In text classification, the attention layer queries the text embedding ($q=h_\ast$) and use the token contextualized vectors as both key and value ($k = v = h$). The prediction is made on $c$. In NLI, we use text embeddings of the premise or hypothesis as query ($q=\bar{h}_\ast$) and keys/values are bi-LSTM representations from the opposite sentence ($k_i=v_i=h_i$). As a result, we obtain two context vectors, $c_{p}$ for the premise and $c_{h}$ for the hypothesis, which are concatenated $[c_p \oplus c_h]$ for prediction.

\section{Constraints on the Objective Function}
\label{sec:objective}

We propose to control the behavior of the attention layer to improve its plausibility by extending the loss function to include a term on the attention
\begin{equation} 
    \label{eq:loss_general}
    \mathcal{L}(y, \hat{y}, \hat{\alpha}) = \mathcal{L}_c(y, \hat{y}) + \lambda \; \mathcal{L}_a(\hat{\alpha})
\end{equation}
where we combine the classification loss $L_c$ (cross-entropy loss) with a constraint on attention map $\mathcal{L}_a(\hat{\alpha})$ weighted by $\lambda \in [0,1]$. We detail hereunder the different forms for $\mathcal{L}_a$ in three approaches.

\subsection{Sparsity Regularization}
\label{sec:objective:regul}

The sparsity constraint can be expressed in many different ways, which have different but marginal effects on convergence speed, or on the resulting explanation~\cite{mohankumar_TransparentExplainableAttention_2020,jia_ARNORAttentionRegularization_2019}. Shannon entropy offers a straightforward yet effective method to measure sparsity, where high entropy values indicate uniform weight distributions and low values indicate sparse ones. We incorporate Shannon entropy as a loss function, defined as
\begin{equation} \label{eq:loss_reg}
    \mathcal{L}_{a}(\hat{\alpha}) = - \sum_{i=1}^{L} \hat{\alpha}_i \log_{L}(\hat{\alpha}_i) \enspace .
\end{equation}

\subsection{Supervision from Reference Annotation}
\label{sec:objective:supervise}

A difficulty in supervising attention layers with a reference annotation is that attention weights and reference annotations are conceptually of different nature. The former are weights such that $\sum\hat{\alpha}_i = 1$ while the latter are binary indicators of whether a token is useful for a plausible explanation or not. Contrary to \cite{nguyen_WhoKilledPolice_2018}, supervision directly on the attention map $\hat{\alpha}_i$ did not work out in practice in the models and tasks that we consider. Thus , we propose instead to supervise on $\hat{\beta}_i = \mbox{sigmoid}(\hat{a}_i)$(similar to logistic attention \cite{martins_SoftmaxSparsemaxSparse_2016}), where $\hat{a}_i$ are the attention before the softmax. Due to the sparsity in the human annotation $\alpha$, the traditional loss function would put too much emphasis on non-annotated tokens so we rather use the Jaccard loss function from \cite{duque-arias_PowerJaccardLosses_2021} to avoid this bias, i.e.,
\begin{equation} 
    \mathcal{L}_{a} (\hat{\beta}, \alpha) = \frac{\hat{\beta}^T \alpha}{\displaystyle \sum_i^{L} \hat{\beta}_i + \sum_i^{L} \alpha_i - \hat{\beta}^T \alpha} \enspace .
    \label{eq:loss_supervision}
\end{equation}
Note that $\hat{\beta}$ is only used in the loss function, $c$ is still computed based on softmax attention map $\hat{\alpha}$.

\subsection{Semi-supervision from Heuristics}
\label{sec:objective:semi}

Supervising with the reference human annotation aims at demonstrating whether supervision can be used to improve plausibility or not in an ideal scenario. This is however not realistic as human annotations are seldom available for this task and are costly to obtain. We thus investigate semi-supervision with annotations generated by simple heuristic rules. Indeed, \cite{nguyen_StudyPlausibilityAttention_2021} show that a simple heuristic attention map exploiting part-of-speech (POS) tags offers decent plausibility in the NLI task. 
The heuristic builds on the observation that verbs, nouns and adjectives (save for those in a small shortlist, such as auxiliary verbs) account for a fair amount of the tokens deemed as informative by human annotators\footnote{POS tags were detected with spaCy using the \textit{en\_core\_web\_sm} pipeline, which claims an accuracy of 97.2\,\% in POS tagging.}. In the e-SNLI dataset, 73.42\,\% of the tokens in the human annotation fall in this category. To a slightly lesser extent, this is also observed on the HateXPlain and Yelp-Hat datasets used of text classification with more than 53\,\% of the annotated tokens in this category.

We construct the heuristic map $\tilde{\alpha} = [\tilde{\alpha}_i]_{i=1}^{L}$ such that $\tilde{\alpha}_i = 0$ for tokens that are not nouns, verbs, adjectives, or stop-words, and reweight the remaining tokens based on the task. For classification tasks, the weight is the frequency of the token in the reference annotation. For NLI task, the weight is the sum of cosine similarities between the token and all tokens in the other sentence, applied equally to premise and hypothesis. Finally in all tasks, the heuristic map $\tilde{\alpha}$ is renormalized on a per-sentence basis, which transformed it into a probability vector. The Kullback-Leibler divergence 
\begin{equation} 
    \label{eq:loss_semisupervision}
    \mathcal{L}_{a}(\hat{\alpha}, \tilde{\alpha}) = \tilde{\alpha} \times [ log(\tilde{\alpha}) - log(\hat{\alpha}) ]
\end{equation}
is used to measure the loss between two probability vectors, $\tilde{\alpha}$ and $\hat{\alpha}$:

Our proposed heuristic for text classification has a limit as it indirectly relies on human annotations to weight each token, but one could make use of semantic lexicons such as SentiWordNet or VerbNet to craft heuristic weights for noun, verb, and adjective tokens. 

\section{Implementation and Training Parameters}
\label{sec:exp}

To ensure consistency, the text data are pre-processed by tokenizing, lemmatizing, and lowercasing using the \textit{spaCy} library. A unique vocabulary was generated for each dataset using the training set. All models reported in this study were initialized with the same GloVE embeddings (glove.42B.300d) and utilized ReLu activation functions with a softmax at the output of the classifier. The training settings were kept at their default configurations, including a learning rate of $lr=1e-3$ and a stabilizer of $\epsilon = 1e{-8}$, as per community standards. To account for model variability, all runs were repeated three times.

Regarding evaluation, assessing the plausibility of attention maps faces three challenges: (1) the attention weights are continuous, (2) the magnitude of its values depends on the sentence length and (3) only a few tokens are highlighted (class imbalance). To address these challenges, we apply a min-max scaler on the attention map and use the Area Under Recall/Precision curve (AURPC) as proposed in ERASER~\cite{deyoung_ERASERBenchmarkEvaluate_2020} to measure how close it is to human annotation. Additionally, we report Recall and Specificity by applying a threshold of 0.5 (the value is chosen following the ERASER benchmark~\cite{deyoung_ERASERBenchmarkEvaluate_2020}) for further insights.

Code to reproduce all experiments is available at 

\noindent \href{https://github.com/Kihansi95/Linkmedia_AttentionPlausibilityByConstraint.git}{https://github.com/Kihansi95/Linkmedia\_AttentionPlausibilityByConstraint}.

\section{Experimental Results}
\label{sec:result}

\begin{figure}[!htb]
    \vspace{-1em}
    \centering
    \includegraphics[width=\textwidth]{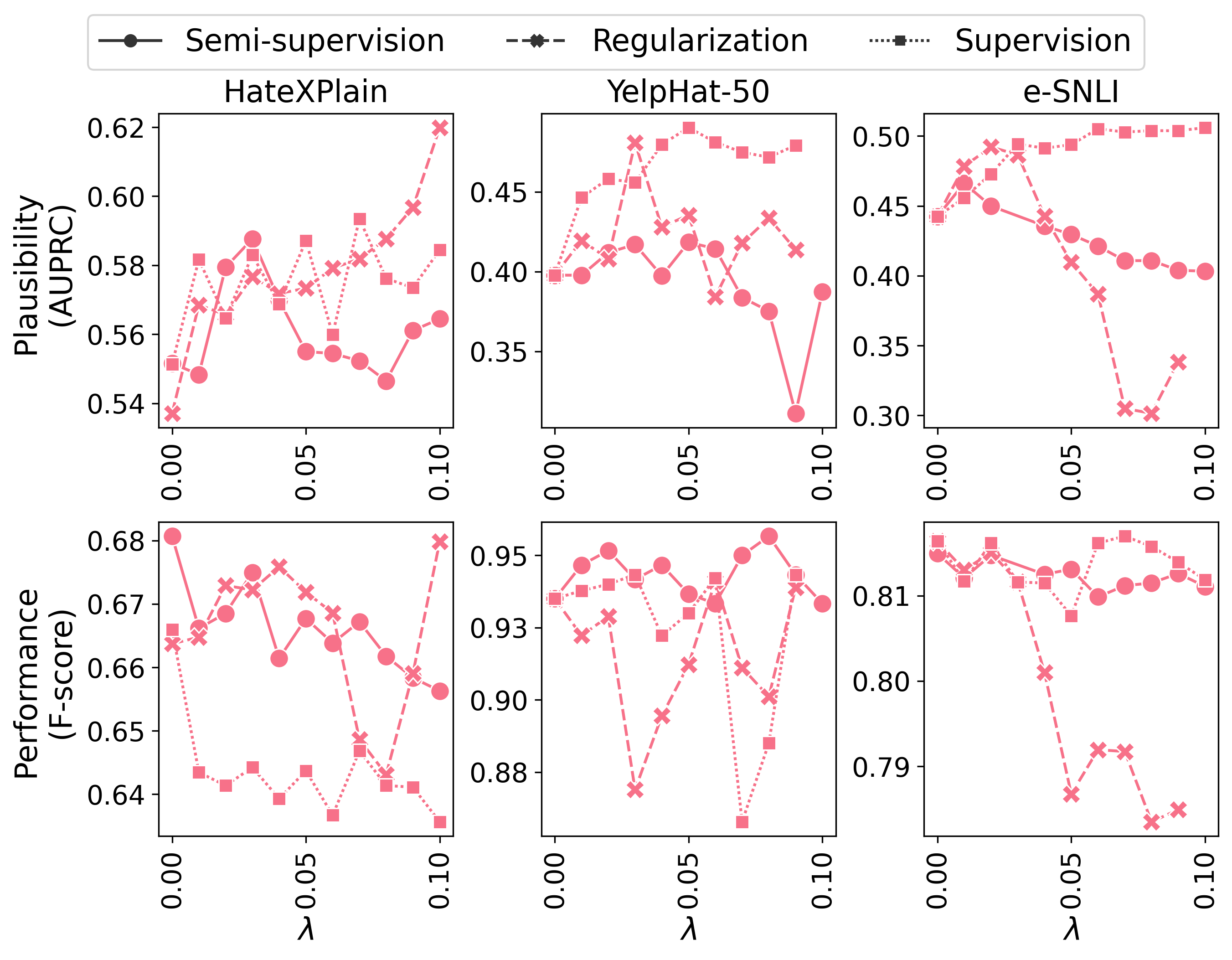}
    \caption{Plausibility (AUPRC, top) and performance (F-score, bottom) on HateXPlain, YelpHat-50, and e-SNLI.}
    \vspace{-2em}
    \label{fig:fscore_auprc}
\end{figure}

Firstly, the study investigates whether enhancing the plausibility of attention map is feasible without compromising classification performance. To answer this question, we evaluate three methods, namely, semi-supervision (solid line with round marker), regularization (dashed line with X marker), and supervision (dotted line with square marker) based on plausibility (AUPRC) and task performance (F-score) as reported in Figure~\ref{fig:fscore_auprc}. The figure showcases the evolution of the two metrics across three datasets under $\lambda$ values ranging from $[0,0.1]$ in a single bi-LSTM contextualization layer setting.

Across all three datasets, supervision and regularization show a consistent improvement in plausibility while preserving the classification performance. Although the semi-supervision shows little effect in HateXPlain, the technique shows improvement in YelpHat and e-SNLI. This suggests that effectiveness of the semi-supervision strategy in general depends on the specific characteristics of the input data, and users cannot always rely on it for improved performance. In fact, the heuristic map for HateXplain is sorely based on most highlighted words by annotators, different from YelpHat dataset where by considering sentiment words can be sufficient for explanations, explanation in HateXPlain depends highly on the context. In many cases, the same words could indicate either a hateful or a non-hateful meaning. The regularization, on the other hand, is very sensitive to $\lambda$, with performance getting hurt rapidly while semi-supervision offers a more stable solution. Notice that supervision in e-SNLI leads to a loss of performance, due to the artifact in annotation instruction~\cite{gururangan_AnnotationArtifactsNatural_2018}.

\begin{figure}[tbhp]
     \centering
     \begin{subfigure}[b]{\textwidth}
         \centering
         \includegraphics[width=\textwidth]{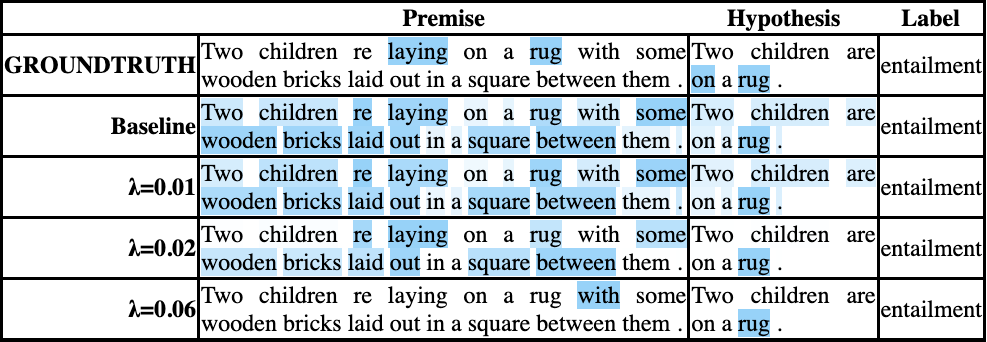}
         \caption{regularization of attention}
         \label{fig:example_esnli_regularization_entailment}
     \end{subfigure}
     \begin{subfigure}[b]{\textwidth}
         \centering
         \includegraphics[width=\textwidth]{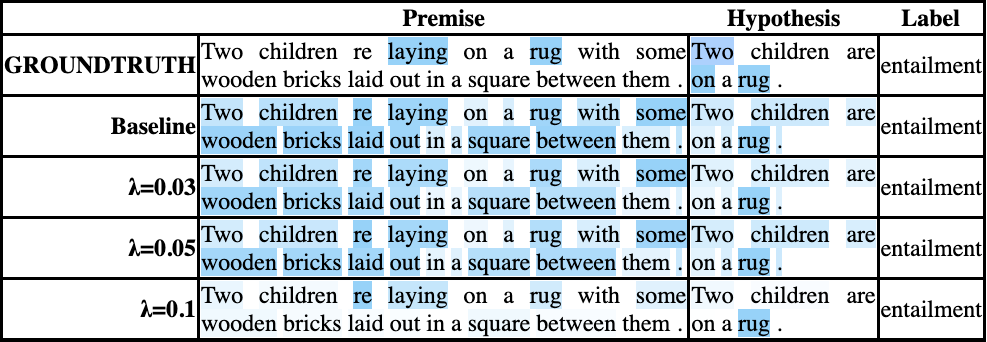}
         \caption{supervision of attention}
         \label{fig:example_esnli_supervision_entailment}
     \end{subfigure}
     \begin{subfigure}[b]{\textwidth}
         \centering
         \includegraphics[width=\textwidth]{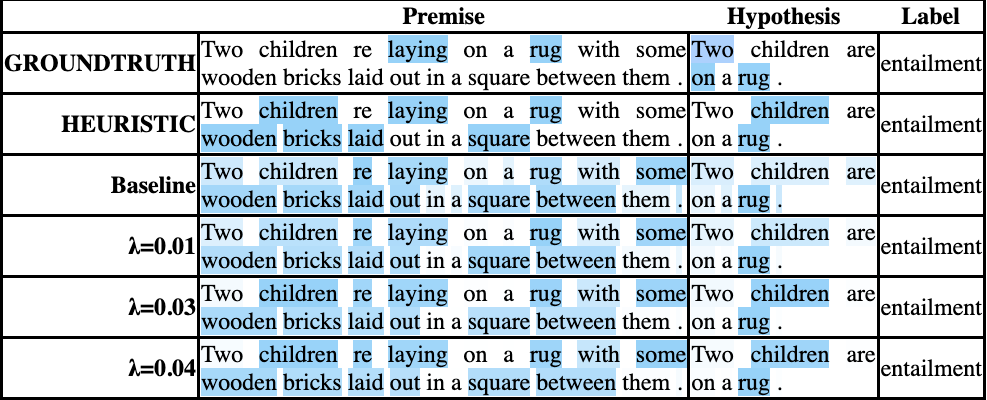}
         \caption{semi-supervision of attention}
         \label{fig:example_esnli_semisupervision_entailment}
     \end{subfigure} 
     \caption{Examples of attention maps on one of the e-SNLI entailment pair.}
     \label{fig:example}
     \vspace{-2.0em}
\end{figure}

The impact of each constraint on attention maps in the NLI task is shown in Figure~\ref{fig:example}. When regularization is strengthened ($\lambda$ increases), the attention maps progressively delete words that were initially highlighted in the baseline ($\lambda = 0$), which is the intended effect of regularization. However, when $\lambda$ surpasses a certain threshold, attention maps become too concentrated on a few words, resulting in less plausible explanations. For instance, in Figure~\ref{fig:example_esnli_regularization_entailment}, when $\lambda=0.06$, the attention maps focus on only one word in each sentence ("with" in premise and "rug" in hypothesis), which renders the explanation implausible and negatively impacts performance (as seen in Figure~\ref{fig:fscore_auprc}, where the F-score drops from 0.815 to around 0.793).

In the case of supervision, attention maps gradually delete words from the baseline model, resulting in more plausible explanations that match the words highlighted by annotators. The constraint, however, does not ensure complete alignment with human annotations as shown in Figure~\ref{fig:example_esnli_supervision_entailment}, where the attention map of $\lambda=0.1$ does not select the words "two" and "on" to explain in hypothesis. In addition, "$\lambda=0.1$" assigns 10\% of the learning loss to make attention maps closer to human annotations, while 90\% of the weight focuses on achieving good classification performance. This limitation means that words picked by human annotation may not all be necessary to model for prediction. As in Figure~\ref{fig:fscore_auprc}, the attention maps cannot be constrained to be more similar to human annotations beyond $\lambda = 0.06$.

In semi-supervision, attention maps tend to keep words obtained from heuristic maps and do not impact performance. For instance, in the hypothesis attention map when $\lambda=0.04$ (Figure~\ref{fig:example_esnli_semisupervision_entailment}), the constraint deletes the words "two", "are", "on" and enforces attention values on "children" to match the heuristic map.

\begin{figure}[tb]
    \vspace{-1em}
    \centering
    \includegraphics[width=\textwidth]{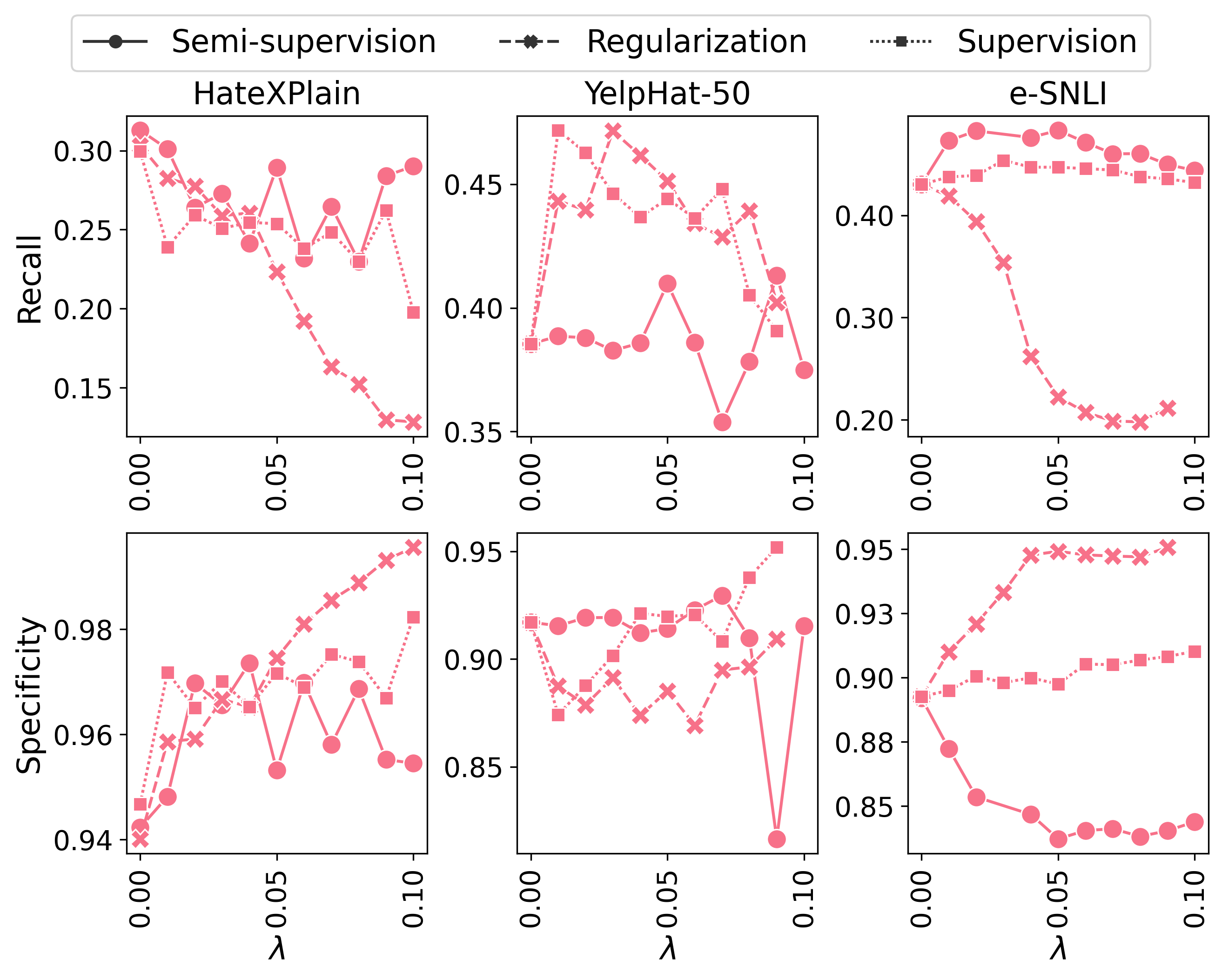}
    \caption{Recall (top) and specificity (bottom) of attention map against annotation.}
    \vspace{-2em}
    \label{fig:precision_specificity}
\end{figure}

To confirm the observation, we report in Figure~\ref{fig:precision_specificity} recall and specificity of attention maps as a function of $\lambda$: while regularization encourages the selection of true positives (increase in recall), it tends to ignore some plausible words as indicated by the drop in specificity (we have more false negatives), as shown in e-SNLI and HateXPlain. This leads to a more conservative model that prefers to drop some words than highlight words that are not plausible. With supervision, the model does the opposite and highlights more correct words by taking the risk of selecting more non-plausible words, thus increasing the false positive rate.

\begin{figure}[htbp]
    \centering
    \begin{subfigure}{\textwidth}  
    \includegraphics[width=\textwidth]{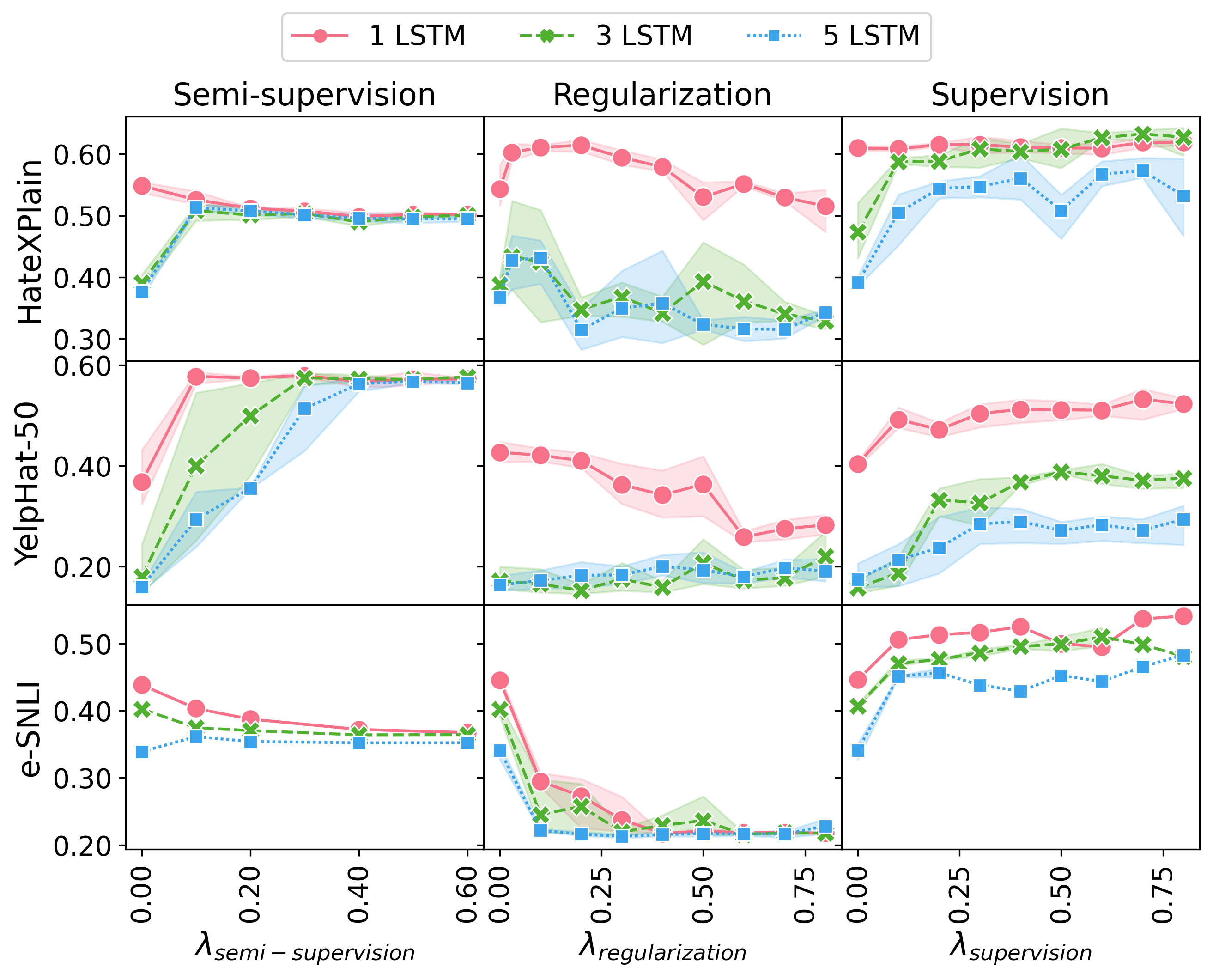}
     \caption{\label{fig:summary_auprc} The plausibility (in AUPRC) of the attention map. } 
    \end{subfigure} 
        
    \begin{subfigure}{\textwidth}  
    \includegraphics[width=.99\textwidth]{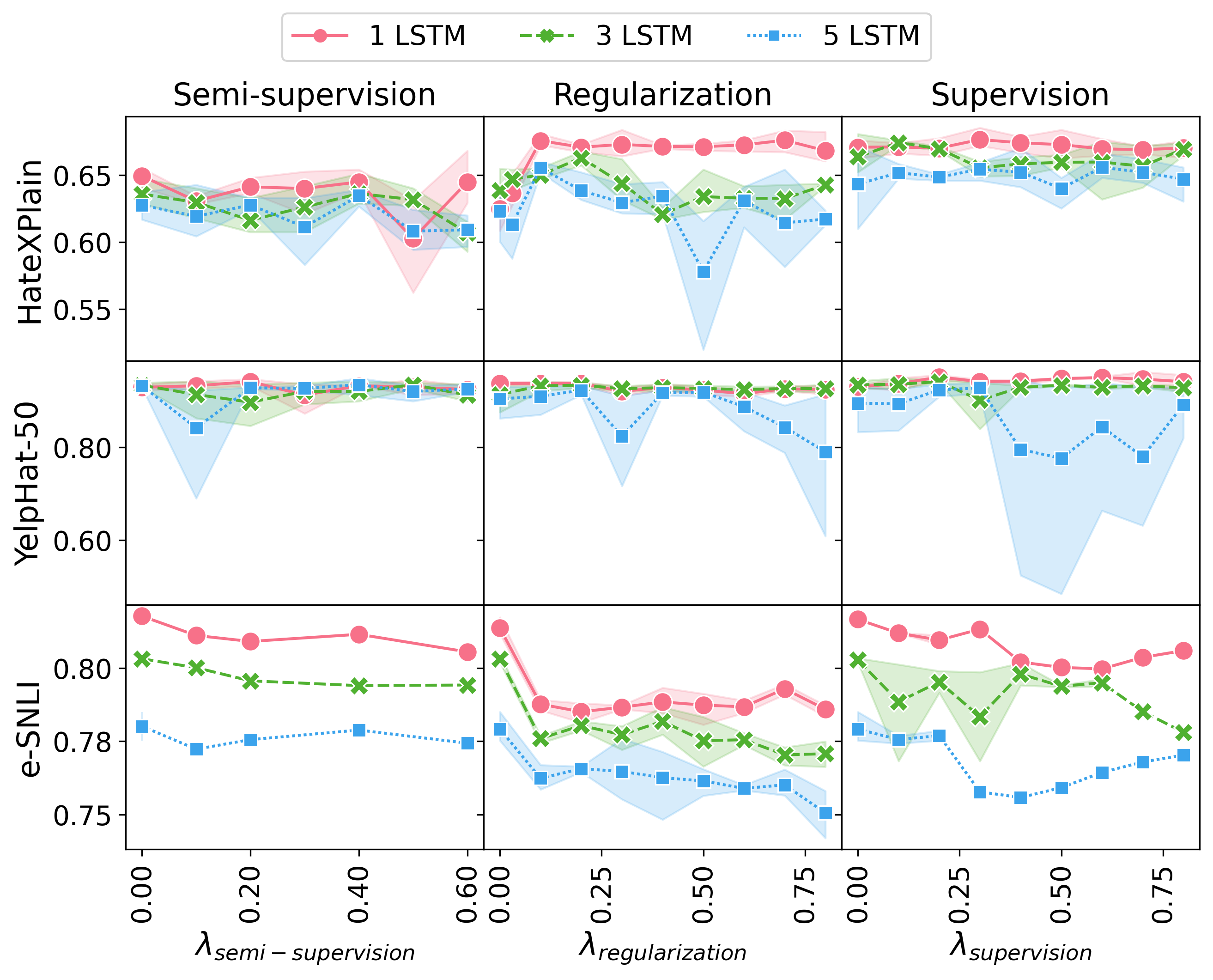}
    \caption{\label{fig:summary_fscore} The task performance (F-Score).}
    \end{subfigure} 

    \caption{Plausibility and performance in 3 datasets, for 3 techniques, for 3 settings.}
    \label{fig:summary}
    \vspace{-2.0em}
\end{figure}

We further study the impact of the LSTM-based contextualization on plausibility. By stacking multiple layers of contextualization, a more semantically meaningful (or deeper) representation of each token can be obtained but also results in a uniform attention map across the entire sentence  \cite{ethayarajh_HowContextualAre_2019,fosse_EtudeStatistiquePlongements_2022}. As regularization and semi-supervision can remove words from the attention map and make it sparse, we explore their potential to overcome the limitation in the deep contextualized model. Figure~\ref{fig:summary_auprc} reports results on the three tasks with one layer (in red (bullets)), three layers (green (crosses)), and five layers (blue (squares)) of bi-LSTMs contextualization, considering the three attention regularization strategies. Note that the scale of $\lambda$ is different in each dataset.

The effect of regularization depends on the task. In easy tasks (YelpHat), regularization does not yield improvement in plausibility. Surprisingly, the semi-supervision can actually improve in the case of the YelpHat corpus, overreaching supervision. In fact, the model's plausibility converges to the AURPC of the heuristic map (0.6546 in YelpHat-50 and 0.5224 for HateXPlain). Although semi-supervision can offer a stable solution in classification tasks, its utility in complex tasks such as NLI and HateXPlain requires careful design. Finally, deeper contextualization with several bi-LSTM layers  makes it harder to obtain a plausible attention map, no matter the technique. This suggests that the contextualization by selectively keeping important features for classification suppressed other information that allow the attention layer to distinguish input tokens between them.

\section{Conclusion}
\label{sec:conclu}

In this work, we compared three approaches to improve the plausibility of attention maps on top of RNN encoders at no extra cost, by adding an attention loss function to the classification loss. Regularization of the attention layer with an entropy criterion limits the words attended to in the model, marginally improving plausibility but risking the deletion of too many plausible tokens or focusing on the wrong ones. Supervision by human annotation encourages attention to focus on words it would not naturally attend to, but it may negatively impact the model's performance depending on the quality and peculiarity of the annotation. Semi-supervision by a heuristic annotation of plausible tokens offers a valuable compromise by improving plausibility without sacrificing performance, but it is limited by the plausibility of the heuristic annotation. We show that the techniques for enforcing plausibility have a lesser impact than the depth of the contextualization with a bi-LSTM encoder. The plausibility of a model decreases with the number of bi-LSTM layers as model performance improves, regardless of attention regularization, suggesting that plausibility from attention in deep transformer-based models remains doubtful. This orients our future efforts to focus on creating an appropriate contextualized vector space that retains enough information to explain the model's decision for humans through contextualization layers.



%
%

%
%
%
\bibliographystyle{splncs04}
\bibliography{2023_NLDB}

\begin{thebibliography}{10}
\providecommand{\url}[1]{\texttt{#1}}
\providecommand{\urlprefix}{URL }
\providecommand{\doi}[1]{https://doi.org/#1}

\bibitem{bai_WhyAttentionsMay_2021}
Bai, B., Liang, J., Zhang, G., Li, H., Bai, K., Wang, F.: Why attentions may
  not be interpretable? In: Proceedings of the 27th {{ACM SIGKDD}} Conference
  on Knowledge Discovery and Data Mining (2021)

\bibitem{bastings_ElephantInterpretabilityRoom_2020}
Bastings, J., Filippova, K.: The elephant in the interpretability room: {{Why}}
  use attention as explanation when we have saliency methods? In: Proceedings
  of the {{Third BlackboxNLP Workshop}} on {{Analyzing}} and {{Interpreting
  Neural Networks}} for {{NLP}} (2020)

\bibitem{bibal_AttentionExplanationIntroduction_2022}
Bibal, A., Cardon, R., Alfter, D., Wilkens, R., Wang, X., Fran{\c c}ois, T.,
  Watrin, P.: Is {{Attention Explanation}}? {{An Introduction}} to the
  {{Debate}}. In: Proceedings of the 60th {{Annual Meeting}} of the
  {{Association}} for {{Computational Linguistics}} (2022)

\bibitem{camburu_ESNLINaturalLanguage_2018}
Camburu, O.M., Rockt{\"a}schel, T., Lukasiewicz, T., Blunsom, P.: E-{{SNLI}}:
  {{Natural Language Inference}} with {{Natural Language Explanations}}. In:
  Proceedings of the 32nd Annual Conference on Neural Information Processing
  Systems (2018)

\bibitem{chen_CanRationalizationImprove_2022}
Chen, H., He, J., Narasimhan, K., Chen, D.: Can {{Rationalization Improve
  Robustness}}? In: Proceedings of the 2022 {{Conference}} of the {{North
  American Chapter}} of the {{Association}} for {{Computational Linguistics}}:
  {{Human Language Technologies}} (2022)

\bibitem{deyoung_ERASERBenchmarkEvaluate_2020}
DeYoung, J., Jain, S., Rajani, N.F., Lehman, E., Xiong, C., Socher, R.,
  Wallace, B.C.: {{ERASER}}: {{A Benchmark}} to {{Evaluate Rationalized NLP
  Models}}. In: Proceedings of the 58th {{Annual Meeting}} of the
  {{Association}} for {{Computational Linguistics}}. {Association for
  Computational Linguistics} (2020)

\bibitem{duque-arias_PowerJaccardLosses_2021}
{Duque-Arias}, D., {Velasco-Forero}, S., Deschaud, J.E., Goulette, F., Serna,
  A., Decenci{\`e}re, E., Marcotegui, B.: On power {{Jaccard}} losses for
  semantic segmentation. In: 16th International Conference on Computer Vision
  Theory and Applications (2021)

\bibitem{ethayarajh_HowContextualAre_2019}
Ethayarajh, K.: How {{Contextual}} are {{Contextualized Word Representations}}?
  {{Comparing}} the {{Geometry}} of {{BERT}}, {{ELMo}}, and {{GPT-2
  Embeddings}}. In: Proceedings of the 2019 {{Conference}} on {{Empirical
  Methods}} in {{Natural Language Processing}} and the 9th {{International
  Joint Conference}} on {{Natural Language Processing}} (2019)

\bibitem{fosse_EtudeStatistiquePlongements_2022}
Fosse, L., Nguyen, D.H., S{\'e}billot, P., Gravier, G.: {Une \'etude
  statistique des plongements dans les mod\`eles transformers pour le
  fran\c{c}ais}. In: 29th Conference Traitement Automatique des Langues
  Naturelles (2022)

\bibitem{gururangan_AnnotationArtifactsNatural_2018}
Gururangan, S., Swayamdipta, S., Levy, O., Schwartz, R., Bowman, S., Smith,
  N.A.: Annotation {{Artifacts}} in {{Natural Language Inference Data}}. In:
  Proceedings of the 2018 {{Conference}} of the {{North American Chapter}} of
  the {{Association}} for {{Computational Linguistics}} (2018)

\bibitem{jacovi_FaithfullyInterpretableNLP_2020}
Jacovi, A., Goldberg, Y.: Towards {{Faithfully Interpretable NLP Systems}}:
  {{How Should We Define}} and {{Evaluate Faithfulness}}? In: Proceedings of
  the 58th {{Annual Meeting}} of the {{Association}} for {{Computational
  Linguistics}} (2020)

\bibitem{jain_AttentionNotExplanation_2019}
Jain, S., Wallace, B.C.: Attention is not {{Explanation}}. In: Proceedings of
  the 2019 {{Conference}} of the {{North American Chapter}} of the
  {{Association}} for {{Computational Linguistics}}. {Minneapolis, Minnesota}
  (2019)

\bibitem{jia_ARNORAttentionRegularization_2019}
Jia, W., Dai, D., Xiao, X., Wu, H.: {{ARNOR}}: {{Attention Regularization}}
  based {{Noise Reduction}} for {{Distant Supervision Relation
  Classification}}. In: Proceedings of the 57th {{Annual Meeting}} of the
  {{Association}} for {{Computational Linguistics}} (2019)

\bibitem{lehman_InferringWhichMedical_2019}
Lehman, E., DeYoung, J., Barzilay, R., Wallace, B.C.: Inferring {{Which Medical
  Treatments Work}} from {{Reports}} of {{Clinical Trials}}. In: Proceedings of
  the 2019 {{Conference}} of the {{North American Chapter}} of the
  {{Association}} for {{Computational Linguistics}} (2019)

\bibitem{luong_EffectiveApproachesAttentionbased_2015}
Luong, T., Pham, H., Manning, C.D.: Effective {{Approaches}} to
  {{Attention-based Neural Machine Translation}}. In: Proceedings of the
  {{Conference}} on {{Empirical Methods}} in {{Natural Language Processing}}
  (2015)

\bibitem{martins_SoftmaxSparsemaxSparse_2016}
Martins, A., Astudillo, R.: From softmax to sparsemax: {{A}} sparse model of
  attention and multi-label classification. In: Proceedings of the 33rd
  International Conference on Machine Learning (2016)

\bibitem{mathew_HateXplainBenchmarkDataset_2021}
Mathew, B., Saha, P., Yimam, S.M., Biemann, C., Goyal, P., Mukherjee, A.:
  {{HateXplain}}: {{A Benchmark Dataset}} for {{Explainable Hate Speech
  Detection}}. In: Proceedings of the AAAI Conference on Artificial
  Intelligence (2021)

\bibitem{mcguire_SentimentAnalysisCognitive_2021}
McGuire, E.S., Tomuro, N.: Sentiment {{Analysis}} with {{Cognitive Attention
  Supervision}}. Proceedings of the Canadian Conference on Artificial
  Intelligence  (2021)

\bibitem{mohankumar_TransparentExplainableAttention_2020}
Mohankumar, A.K., Nema, P., Narasimhan, S., Khapra, M.M., Srinivasan, B.V.,
  Ravindran, B.: Towards {{Transparent}} and {{Explainable Attention Models}}.
  In: Proceedings of the 58th {{Annual Meeting}} of the {{Association}} for
  {{Computational Linguistics}} (2020)

\bibitem{nguyen_StudyPlausibilityAttention_2021}
Nguyen, D.H., Gravier, G., S{\'e}billot, P.: A {{Study}} of the
  {{Plausibility}} of {{Attention}} between {{RNN Encoders}} in {{Natural
  Language Inference}}. In: 20th {{IEEE International Conference}} on {{Machine
  Learning}} and {{Applications}} (2021)

\bibitem{nguyen_FiltrageRegularisationPour_2022}
Nguyen, D.H., Gravier, G., S{\'e}billot, P.: Filtrage et r\'egularisation pour
  am\'eliorer la plausibilit\'e des poids d'attention dans la t\^ache
  d'inf\'erence en langue naturelle. In: Traitement {{Automatique}} Des
  {{Langues Naturelles}} (2022)

\bibitem{nguyen_WhoKilledPolice_2018}
Nguyen, M., Nguyen, T.H.: Who is {{Killed}} by {{Police}}: {{Introducing
  Supervised Attention}} for {{Hierarchical LSTMs}}. In: Proceedings of the
  27th {{International Conference}} on {{Computational Linguistics}} (2018)

\bibitem{ousidhoum_ProbingToxicContent_2021}
Ousidhoum, N., Zhao, X., Fang, T., Song, Y., Yeung, D.Y.: Probing {{Toxic
  Content}} in {{Large Pre-Trained Language Models}}. In: Proceedings of the
  59th {{Annual Meeting}} of the {{Association}} for {{Computational
  Linguistics}} and the 11th {{International Joint Conference}} on {{Natural
  Language Processing}} (2021)

\bibitem{paranjape_InformationBottleneckApproach_2020}
Paranjape, B., Joshi, M., Thickstun, J., Hajishirzi, H., Zettlemoyer, L.: An
  {{Information Bottleneck Approach}} for {{Controlling Conciseness}} in
  {{Rationale Extraction}}. In: Proceedings of the 2020 {{Conference}} on
  {{Empirical Methods}} in {{Natural Language Processing}} (2020)

\bibitem{pruthi_LearningDeceiveAttentionBased_2020}
Pruthi, D., Gupta, M., Dhingra, B., Neubig, G., Lipton, Z.C.: Learning to
  {{Deceive}} with {{Attention-Based Explanations}}. In: Proceedings of the
  58th {{Annual Meeting}} of the {{Association}} for {{Computational
  Linguistics}} (2020)

\bibitem{sen_HumanAttentionMaps_2020a}
Sen, C., Hartvigsen, T., Yin, B., Kong, X., Rundensteiner, E.: Human
  {{Attention Maps}} for {{Text Classification}}: {{Do Humans}} and {{Neural
  Networks Focus}} on the {{Same Words}}? In: Proceedings of the 58th {{Annual
  Meeting}} of the {{Association}} for {{Computational Linguistics}} (2020)

\bibitem{serrano_AttentionInterpretable_2019}
Serrano, S., Smith, N.A.: Is {{Attention Interpretable}}? In: Proceedings of
  the 57th {{Annual Meeting}} of the {{Association}} for {{Computational
  Linguistics}} (2019)

\bibitem{sun_UnderstandingAttentionText_2020}
Sun, X., Lu, W.: Understanding {{Attention}} for {{Text Classification}}. In:
  Proceedings of the 58th {{Annual Meeting}} of the {{Association}} for
  {{Computational Linguistics}} (2020)

\bibitem{vashishth_AttentionInterpretabilityNLP_2019}
Vashishth, S., Upadhyay, S., Tomar, G.S., Faruqui, M.: Attention
  {{Interpretability Across NLP Tasks}}. CoRR  (2019)

\bibitem{wiegreffe_TeachMeExplain_2021}
Wiegreffe, S., Marasovi{\'c}, A.: Teach {{Me}} to {{Explain}}: {{A Review}} of
  {{Datasets}} for {{Explainable Natural Language Processing}}. In: Proceedings
  of the Neural Information Processing Systems Track on Datasets and Benchmark.
  vol.~1 (2021)

\bibitem{wiegreffe_AttentionNotNot_2019}
Wiegreffe, S., Pinter, Y.: Attention is not not {{Explanation}}. In:
  Proceedings of the 2019 {{Conference}} on {{Empirical Methods}} in {{Natural
  Language Processing}} and the 9th {{International Joint Conference}} on
  {{Natural Language Processing}} (2019)

\end{thebibliography}
\end{document}